\documentclass[11pt]{article}

\usepackage{fancyhdr}

\usepackage{geometry}

\usepackage{thumbpdf,lmodern}
\usepackage{bm}
\usepackage{bbm}
\usepackage{float}
\usepackage{amsthm}
\usepackage{amssymb}
\usepackage{subfig}
\usepackage[round]{natbib}
\usepackage{enumerate}
\usepackage{longtable}
\usepackage{mathtools,cancel}
\usepackage{url}
\usepackage[normalem]{ulem}

\usepackage[ruled,vlined,linesnumbered,lined,boxed,commentsnumbered]{algorithm2e}
\usepackage{tikz}
\usetikzlibrary{positioning}
\usetikzlibrary{graphs} 
\usetikzlibrary[graphs]
\usepackage{multirow}
\usepackage{lscape}
\usepackage[english]{babel}
\usepackage{amsmath}

\usepackage{tikz}
\usetikzlibrary{positioning}

\RequirePackage[OT1]{fontenc}
\RequirePackage{amsthm,amsmath}

\usepackage{authblk}

\SetKwInput{KwInput}{Input}
\SetKwInput{KwOutput}{Output}

\usepackage[linesnumbered,ruled,vlined]{algorithm2e}

\usepackage{framed}
\usepackage{xcolor}

\renewcommand{\Re}{\mathbb{R}}

\newcommand{\vat}{\mathbb{E}}
\newcommand{\var}{\mathbb{V}\!\mathrm{ar}}

\newcommand{\bc}{\begin{center}}
	\newcommand{\ec}{\end{center}}

\newcommand{\bit}{\begin{itemize}}
	\newcommand{\eit}{\end{itemize}}

\newcommand{\be}{\begin{eqnarray*}}
	\newcommand{\ee}{\end{eqnarray*}}
\newcommand{\ben}{\begin{eqnarray}}
	\newcommand{\een}{\end{eqnarray}}

\newcommand{\g}{\,\vert\,}

\newcommand{\D}{\mathcal{D}}

\newcommand{\N}{\mathcal{N}}

\newcommand{\pa}{\mathrm{pa}}
\newcommand{\fa}{\mathrm{fa}}

\newcommand{\bzero}{\bm{0}}

\newcommand{\bA}{\bm{A}}

\newcommand{\bD}{\bm{D}}

\newcommand{\bI}{\bm{I}}

\newcommand{\bL}{\bm{L}}

\newcommand{\bS}{\bm{S}}

\newcommand{\bU}{\bm{U}}
\newcommand{\bX}{\bm{X}}

\newcommand{\bx}{\bm{x}}

\newcommand{\bSigma}{\bm{\Sigma}}

\newcommand{\bOmega}{\bm{\Omega}}

\begin{document}

\title{\textbf{\texttt{BCDAG}}: An \textbf{\texttt{R}} package for Bayesian structure and Causal learning of Gaussian DAGs}
\author[1]{Federico Castelletti \thanks{federico.castelletti@unicatt.it}}
\author[2]{Alessandro Mascaro \thanks{a.mascaro3@campus.unimib.it}}
\affil[1]{Department of Statistical Sciences, Universit\`{a} Cattolica del Sacro Cuore, Milan}
\affil[2]{Department of Economics, Management and Statistics, Universit\`{a} degli Studi di Milano-Bicocca, Milan}

\date{}

\maketitle

\begin{abstract}
Directed Acyclic Graphs (DAGs) provide a powerful framework to model causal relationships among variables in multivariate settings; in addition, through the \textit{do-calculus} theory, they allow for the identification and estimation of causal effects between variables also from pure observational data.
In this setting, the process of inferring the DAG structure from the data is referred to as causal structure learning or causal discovery.
We introduce \textbf{\texttt{BCDAG}}, an \textbf{\texttt{R}} package for Bayesian causal discovery and causal effect estimation from Gaussian observational data, implementing the Markov chain Monte Carlo (MCMC) scheme proposed by \cite{Castelletti:Mascaro:2021}. Our implementation scales efficiently with the number of observations and, whenever the DAGs are sufficiently \textit{sparse}, with the number of variables in the dataset.
The package also provides functions for convergence diagnostics and for visualizing and summarizing posterior inference. In this paper, we present the key features of the underlying methodology along with its implementation in \textbf{\texttt{BCDAG}}. We then illustrate the main functions and algorithms on both real and simulated datasets.

\vspace{0.7cm}
\noindent
Keywords: Graphical model, Bayesian structure learning, Causal inference, Markov chain Monte Carlo, \textbf{\texttt{R}}

\end{abstract}

\section[Introduction]{Introduction}
\label{sec:intro}


In the last decades, probabilistic graphical models \citep{Laur:1996} have emerged as a powerful tool
for modelling and inferring \emph{dependence} relations in complex multivariate settings.
Specifically, Directed Acyclic Graphs (DAGs), also called Bayesian networks, adopt a graph-based representation to model a given set of conditional independence statements between variables which defines the DAG \textit{Markov property}.
In addition,
if coupled with the \textit{do-calculus} theory \citep{Pear:2000}, DAGs can be also adopted for causal inference and allow to identify and estimate \emph{causal} relationships between variables.
When DAGs are endowed with causal assumptions, the process of learning their graphical structure is usually referred to as \emph{causal structure learning} or simply \emph{causal discovery}
\citep{peters2017} and often employed in a non-experimental setting, namely when only observational data are available; see \citet{Maathuis:Nandy:Review} for a review.
In this setting, assuming faithfulness and causal sufficiency 
\citep{Spir:Glym:Sche:2000}
it is possible to learn the DAG structure only up to its Markov equivalence class \citep{Ande:etal:1997}, which collects all DAGs having the same Markov property.
However, because equivalent DAGs 
still represent different Structural Causal Models \citep{Pear:2000} (SCMs), a collection of potentially distinct causal effects can be recovered from the estimated Markov equivalence class \citep{Maat:etal:2009}.

The task of causal discovery has been tackled
from different methodological perspectives, and in particular under both the frequentist and Bayesian frameworks. 
A primary distinction among frequentist methods is between \textit{constraint-based} and \textit{score-based} algorithms.
The former include algorithms that recover the DAG-equivalence class
through sequences of
conditional independence tests.
The most popular methods are the PC and Fast Causal Inference (FCI) algorithms \citep{Spir:Glym:Sche:2000},
together with their extensions rankPC and rankFCI \citep{harrisdrton2013}, based on more general (non-parametric) conditional independence tests.
Differently, score-based methods implement a suitable score function
which is maximized over the space of DAGs (or their equivalence classes)
to provide a graph estimate; examples are the Greedy Equivalence Search (GES) algorithm of \cite{chickering2002}
and the HC Bdeu method \citep{russell2002artificial}.
Going beyond this distinction, a variety of hybrid methods,
i.e.~combining features of both the two approaches,
have been proposed; see for instance
\cite{tsamardinos2006}, \cite{solus2021} and
\cite{shimizu2006}, the latter tailored to
non-Gaussian linear structural equation models.
For an extensive review of causal discovery methods
the reader can refer to \citet{heinze2018causal}.

\textbf{\texttt{R}} implementations of frequentist approaches for structure learning are available within a few packages, the most popular being \textbf{\texttt{bnlearn}} and \textbf{\texttt{pcalg}}.
Specifically, \textbf{\texttt{bnlearn}} \citep{scutari2010bnlearn} implements both structure learning and parameter estimation for discrete and Gaussian Bayesian networks. However, it
is not specifically tailored to causal inference since it does not provide estimation of causal effects.
On the other hand, \textbf{\texttt{pcalg}} \citep{kalisch2012pcalg} focuses on causal inference applications; it thus implements various algorithms for causal discovery from observational and experimental data \citep{heinze2018invariant, hauser2012characterization} and causal effect estimation when the DAG is unknown \citep{Maat:etal:2009,nandy2017estimating}.
Outside the \textbf{\texttt{R}} environment, \textbf{\texttt{Python}} implementations of causal discovery and causal inference methodologies are available within the \textbf{\texttt{Causal discovery toolbox}} of  \cite{kalainathan2019causal}, \textbf{\texttt{causal-dag}} \citep{squires2018causal} and \textbf{\texttt{causal-learn}} \citep{cmu2022causal}, the last providing a Python extension of \textbf{\texttt{Tetrad}} \citep{glymour1988tetrad}, the historical \textbf{\texttt{Java}} application for causal discovery.

On the Bayesian side, DAG structure learning has been traditionally tackled as a Bayesian model selection problem;
in this framework the target is represented by the posterior distribution of graph structures which is typically approximated through Markov Chain Monte Carlo (MCMC) methods; see for instance
\citet{Cooper:Herskovits:1992,Ni:et:al:2017,Castelletti:et:al:2018}.
Differently from frequentist methods, Bayesian techniques
provide a coherent quantification of the uncertainty around DAG estimates.
By converse however, they require the elicitation of a suitable parameter prior distribution for each candidate DAG-model.
To this end,
\citet{Heckerman:1995} and \citet{Geig:Heck:2002} proposed an effective procedure which assigns priors to DAG-model parameters \textit{via} a small number of direct assessments and guarantees score equivalence for Markov equivalent DAGs.
Most importantly the two methods, developed for categorical and Gaussian DAG-models respectively, lead to closed-form expressions for the DAG marginal likelihood, which serves as input to \emph{model selection} algorithms based on MCMC schemes. 
More recently, \citet{Ben:Massam:arXiv} and \citet{cao:et:al:2019} introduced multi-shape DAG-Wishart
distributions 
as conjugate priors for Gaussian DAG-models.
Accordingly, their framework also
allows for \emph{posterior inference} on model parameters,
a feature which is also essential for causal effect estimation \citep{Caste:Cons:2021}.

R implementations of Bayesian methods for DAG structure learning are provided by the packages \textbf{\texttt{mcmcabn}} and \textbf{\texttt{BiDAG}} \citep{suter2021bidag}  among a few others.
The former develops structure MCMC algorithms to approximate a posterior distribution over DAGs, and is implemented for both discrete and continuous data.
Differently, \textbf{\texttt{BiDAG}} implements order and partition MCMC algorithms in the hybrid approach of \citet{Kuipers:et:al:2021:arxiv} which has been shown to perform better in terms of convergence as the number of variables increases.
Both packages are tailored to DAG-model selection only and do not provide any implementation of methods for parameter estimation and causal inference.  
To our knowledge, no implementation of Bayesian methods for causal inference based on DAGs is available.

In this context we introduce \textbf{\texttt{BCDAG}}, an \textbf{\texttt{R}} package \citep{R:core:team} for Bayesian structure learning and causal effect estimation from observational Gaussian data.
\textbf{\texttt{BCDAG}} implements an efficient Partial Analytic Structure algorithm \citep{Godsill:2012} to sample from the joint posterior distribution of DAGs and DAG parameters and applies the Bayesian approach of \citet{Castelletti:Mascaro:2021} for causal effect estimation.
Accordingly, our package combines i) structure learning of Gaussian DAGs, ii) posterior inference of DAG-model parameters and iii) estimation of causal effects between variables in the dataset.

The rest of the paper is organized as follows.
In Section \ref{sec:Gaussian:DAGs} we introduce Gaussian DAG-models from a Bayesian perspective. Specifically, we write the likelihood and define suitable prior distributions for both DAG structures and DAG-model parameters.
In the same section we also summarize a few useful results regarding parameter posterior distributions and DAG marginal likelihoods.
We then provide in Section \ref{sec:causal:effects} the definition of causal effect in a Gaussian-DAG setting.
Section \ref{sec:MCMC} presents the main MCMC scheme that we adopt for posterior inference of DAGs and parameters and in turn for causal effect estimation.
Illustrations of the main functions and algorithms are provided throughout the paper and described more extensively in Section \ref{sec:illustrations} on both simulated and real data. Finally, Section \ref{sec:discussion} presents a brief discussion.

\section{Gaussian DAG-models}
\label{sec:Gaussian:DAGs}

Let $\D=(V,E)$ be a Directed Acyclic Graph (DAG), where $V=\{1,\dots,q\}$ is a set of vertices (or nodes) and $E\subseteq V \times V$ a set of edges.
If $(u,v) \in E$ then $\D$ contains the directed edge $u \rightarrow v$. In addition, $\D$ cannot contain cycles, that is paths of the form $u_0 \rightarrow u_1 \rightarrow \dots \rightarrow u_k$ where $u_0 \equiv u_k$. For a given node $v$, if $u\rightarrow v \in E$ we say that $u$ is a \emph{parent} of $v$; conversely $v$ is a \emph{child} of $u$. The set of all parents of $v$ in $\D$ is denoted by $\pa_{\D}(v)$, while the set $\fa_{\D}(v)=v \cup \pa_{\D}(v)$ is called the \textit{family} of $v$ in $\D$. Furthermore, a DAG is \textit{complete} if all its nodes are joined by edges.
Finally, a DAG $\D$ can be uniquely represented through its $(q,q)$ \emph{adjacency matrix} $\bA^{\D}$ such that $\bA^{\D}_{u,v} = 1$ if and only $\D$ contains $u \rightarrow v$ and $0$ otherwise.

A DAG $\D$ encodes a set of conditional independencies between nodes (variables) that can be read-off from the DAG using graphical criteria, such as \textit{d-separation} \citep{Pear:2000}. The resulting set of conditional independencies embedded in $\D$ defines the DAG \textit{Markov} property.

In the next sections we define a Gaussian DAG-model in terms of likelihood and prior distributions for model parameters. Preliminary results and functions needed for the MCMC algorithm presented in Section \ref{sec:MCMC} are also introduced.

\subsection{Likelihood}
\label{subsec:likelihood}

Let $\D=(V,E)$ be a DAG, $(X_1,\dots,X_q)$ a collection of real-valued random variables each associated to a node in $V$.
We assume that the joint density of $(X_1,\dots,X_q)$ belongs to a zero-mean Gaussian DAG-model, namely
\ben
\label{eq:Gaussian:DAG:model}
X_1,\dots,X_q\g\bOmega_{\D}
\sim
\N_q(\bzero,\bOmega_{\D}^{-1}),
\quad \bOmega_{\D} \in \mathcal{P}_{\D},
\een
where $\bOmega_{\D}=\bSigma_{\D}^{-1}$ is the precision (inverse-covariance) matrix, and
$\mathcal{P}_{\D}$ is the space of symmetric positive definite (s.p.d.) precision matrices Markov w.r.t. $\D$. Accordingly, $\bOmega_{\D}$ satisfies the conditional independencies (Markov property) encoded by $\D$.
For the remainder of this section we will assume the DAG fixed and therefore omit the dependence on $\D$ from the DAG-parameter $\bOmega_{\D}$.

\vspace{0.1cm}

An alternative representation of
model \eqref{eq:Gaussian:DAG:model} is given by the allied Structural Equation Model (SEM). Specifically, let
$\bL$ be a $(q,q)$ matrix of coefficients such that for each $(u,v)$-element $\bL_{uv}$ with $u\ne v$, $\bL_{uv}\ne 0$ if and only if $(u,v) \in E$, while $\bL_{uu}=1$ for each $u=1,\dots,q$.
Let also $\bD$ be a $(q,q)$ diagonal matrix with $(u,u)$-element $\bD_{uu}$.
The SEM representation of \eqref{eq:Gaussian:DAG:model} is then
\ben
\label{eq:SEM}
\bL^\top(X_1,\dots,X_q)^\top = \boldsymbol{\varepsilon}, \quad \boldsymbol{\varepsilon} \sim \N_q(\bzero,\bD),
\een
which implies the re-parameterization $\bOmega=\bL\bD^{-1}\bL^{\top}$. The latter equality is sometimes referred to as the \textit{modified Cholesky decomposition} of $\bOmega$ \citep{cao:et:al:2019}
and induces a re-parametrization of $\bOmega = \bSigma^{-1}$
in terms of node-parameters
$\big\{
(\bD_{jj}, \bL_{\prec j\,]} ), \,
j=1,\dots,q
\big\}$, such that
\be
\bL_{ \prec j\,]}=-\bSigma_{ \prec j\, \succ}^{-1}\bSigma_{ \prec j\,]}, \quad
\bD_{jj} = \bSigma_{jj|\pa_{\D}(j)},
\ee
where $\bSigma_{jj|\pa_{\D}(j)} = \bSigma_{jj}-\bSigma_{[\, j\succ} \bSigma^{-1}_{\prec j\succ} \bSigma_{\prec j\,]}$,
$\prec j\,] = \pa_{\D}(j)\times j$, $[\, j\succ\, = j \times \pa_{\D}(j)$, $\prec j\succ\, = \pa_{\D}(j)\times \pa_{\D}(j)$. In particular, parameter $\bD_{jj}$ corresponds to the conditional variance of $X_j$, $\var(X_j\g\bx_{\pa_{\D}(j)})$.
Base on \eqref{eq:SEM}, model \eqref{eq:Gaussian:DAG:model} can be equivalently re-written as
\ben
\label{eq:Gaussian:DAG:Chol:facorization}
f(x_1,\dots,x_q\g \bD, \bL)
=
\prod_{j=1}^{q}d\,\N(x_j\g -\bL_{ \prec j\,]}^\top\bx_{\pa_{\D}(j)}, \bD_{jj}).
\een

Consider now $n$ independent samples $\bx_i=(x_{i,1},\dots,x_{i,q})^\top$, $i=1,\dots,n$,
from \eqref{eq:Gaussian:DAG:Chol:facorization}, and let $\bX$ be the $(n,q)$ data matrix, row-binding of $\bx_i, \dots, \bx_n$.
The likelihood function is then
\ben
\label{eq:likelihood}
\begin{aligned}
	f(\bX \g \bD, \bL)
	&=
	\prod_{i=1}^{n}
	f(x_{i,1},\dots,x_{i,q}\g \bD, \bL)\\
	&=
	\prod_{j=1}^{q}
	d\,\N_n(\bX_j\g -\bX_{\pa_{\D}(j)} \bL_{ \prec j\,]}, \bD_{jj} \bI_n),
\end{aligned}
\een
where $\bX_{A}$ is the $(n,|A|)$ sub-matrix of $\bX$ corresponding to the set $A$ of columns of $\bX$ and $\bI_n$ denotes the $(n,n)$ identity matrix.
We now proceed by assigning a suitable prior distribution to the DAG-dependent parameters $(\bD,\bL)$.

\subsection{DAG-Wishart prior}


Let $\bOmega$ be the precision matrix Markov w.r.t.~to DAG $\D$.
Conditionally on $\D$, we assign a prior to $\bOmega$ through a \emph{DAG-Wishart} prior on $(\bD,\bL)$
with rate hyperparameter $\bU$ (a $q\times q$ s.p.d. matrix)
and shape hyperparameter $\boldsymbol{a}^{\D} = (a_1^{\D},\dots,a_q^{\D})^\top$ \citep{Ben:Massam:arXiv}.
An important feature of the DAG-Wishart distribution is that node-parameters $\big\{
(\bD_{jj}, \bL_{\prec j\,]} ), \,
j=1,\dots,q
\big\}$ are a priori independent with distribution
\ben
\label{eq:prior:cholesky}
\begin{aligned}
	\bD_{jj} \g \D \sim& \,\,\, \textnormal{I-Ga}\left(\frac{1}{2}a_j^{\D},
	\frac{1}{2}\bU_{jj|\pa_{\D}(j)}\right), \\
	\bL_{\prec j\,]}\g\bD_{jj}, \D \sim& \,\,\, \N_{|\pa_{\D}(j)|}\left(-\bU_{\prec j \succ}^{-1}\bU_{\prec j\,]},\bD_{jj}\bU_{\prec j \succ}^{-1}\right),
\end{aligned}
\een
where $\textnormal{I-Ga}(\alpha,\beta)$ stands for an Inverse-Gamma distribution with shape $\alpha>0$ and rate $\beta>0$ having expectation $\beta/(\alpha-1)$ ($\alpha>1$).
Parameters $a_1^{\D},\dots,a_q^{\D}$ are specific to the DAG-model under consideration. The default choice, hereinafter considered,
$a_j^{\D}=a+|\pa_{\D}(j)|-q+1$ $(a>q-1)$
guarantees \textit{compatibility} among prior distributions for Markov equivalent DAGs. In particular, it can be shown that under this choice any two Markov equivalent DAGs are assigned the \textit{same} marginal likelihood; see also Section \ref{sec:marg:likelihood}.
A prior on parameters $(\bD,\bL)$ is then given by
\ben
\label{eq:prior:indep:chol}
p(\bD,\bL\g\D)=\prod_{j=1}^{q}p(\bL_{\prec j\,]}\g\bD_{jj})\,p(\bD_{jj}).
\een
We refer to the resulting prior as the \emph{compatible} DAG-Wishart distribution;
see in particular \citet{Peluso:Consonni:2020} for full details.

\subsection{Sampling from DAG-Wishart distributions}
\label{subsec:DAG-Wishart:sampling}

Given the results summarized in the previous section, Algorithm \ref{alg:DAG-Wishart} implements a direct sampling from a compatible DAG-Wishart distribution.

\begin{algorithm}{
		\SetAlgoLined
		\vspace{0.1cm}
		\KwInput{$n$, $\D$, $a$, $\bU$ \\
			\KwOutput{$n$ draws from a compatible DAG-Wishart distribution with parameters $a,\bU$}\\
			Set $a_j^{\D}=a+|\pa_{\D}(j)|-q+1$ for $j=1,\dots,q$ \\
			\For{$i=1,\dots,n$}{
				Set $\bD^{(i)} = \boldsymbol{0}_q$ a $(q,q)$ null matrix,
				$\bL^{(i)} = \boldsymbol{I}_q$ a $(q,q)$ identity matrix\\
				\For{$j=1,\dots,q$}{
					Randomly draw elements $\big(\bD_{jj}^{(i)}, \bL^{(i)}_{ \prec j\,]}\big)$ from distributions in \eqref{eq:prior:cholesky}
				}
			}
		}
		\Return $\big\{\bD^{(i)},\bL^{(i)}\big\},\dots,\big\{\bD^{(n)},\bL^{(n)}\big\}$
		\caption{Sampling from a DAG-Wishart distribution}
		\label{alg:DAG-Wishart}
	}
\end{algorithm}

Algorithm \ref{alg:DAG-Wishart} is implemented within our package in the function \texttt{rDAGWishart(n, DAG, a, U)}. Arguments of the function are:
\begin{itemize}
	\item \texttt{n}: the number of draws;
	\item \texttt{DAG}: the $(q,q)$ adjacency matrix of DAG $\D$;
	\item \texttt{a}: the common shape hyperparameter of the DAG-Wishart distribution, $a>q-1$;
	\item \texttt{U}: the rate hyperparameter of the DAG-Wishart distribution, a $(q,q)$ s.p.d. matrix. 
\end{itemize}

Consider the following example with $q=4$ variables and a DAG structure corresponding to DAG $\D$ in Figure \ref{fig:DAGs}.

\begin{verbatim}
	q <- 4
	DAG <- matrix(c(0,1,1,0,0,0,0,1,0,0,0,1,0,0,0,0), nrow = q)
	DAG
	     [,1] [,2] [,3] [,4]
	[1,]    0    0    0    0
	[2,]    1    0    0    0
	[3,]    1    0    0    0
	[4,]    0    1    1    0
	
	outDL <- rDAGWishart(n = 1, DAG = DAG, a = q, U = diag(1, q))
	
	outDL$D
	          [,1]      [,2]     [,3]     [,4]
	[1,] 0.9651437 0.0000000 0.000000 0.000000
	[2,] 0.0000000 0.2840032 0.000000 0.000000
	[3,] 0.0000000 0.0000000 1.188965 0.000000
	[4,] 0.0000000 0.0000000 0.000000 5.890211
	
	outDL$L
	          [,1]        [,2]      [,3] [,4]
	[1,]  1.000000  0.00000000  0.000000    0
	[2,]  1.169280  1.00000000  0.000000    0
	[3,] -1.659849  0.00000000  1.000000    0
	[4,]  0.000000 -0.05807009 -1.379419    1
\end{verbatim}

Matrices \texttt{D} and \texttt{L} represent one draw from a compatible DAG-Wishart distribution with parameters $a=q$, $U=\bI_q$ and DAG structure represented by object \texttt{DAG}.

\subsection{DAG-Wishart posterior}
\label{subsec:DAG-Wishart:posterior}

Because of conjugacy of \eqref{eq:prior:cholesky} with the likelihood \eqref{eq:likelihood}, the posterior distribution of $(\bD,\bL)$ given the data $\bX$, $p(\bD,\bL\g\D,\bX)$, is such that for $j=1,\dots,q$
\ben
\label{eq:posterior:chol}
\begin{aligned}
	\bD_{jj} \g \D, \bX \sim& \,\,\, \textnormal{I-Ga}\left( \frac{1}{2}\widetilde a_j^{\D},
	\frac{1}{2}\widetilde\bU_{jj|\pa_{\D}(j)}\right), \\
	\bL_{\prec j\,]}\g\bD_{jj}, \D, \bX \sim& \,\,\, \N_{|\pa_{\D}(j)|}\left(-\widetilde\bU_{\prec j \succ}^{-1}\widetilde\bU_{\prec j\,]},\bD_{jj}\widetilde\bU_{\prec j \succ}^{-1}\right),
\end{aligned}
\een
where
$\widetilde a_j^{\D} = \widetilde{a}+|\pa_{\D}(j)|-q+1$, $\widetilde{a} = a+n$ and
$
\widetilde \bU = \, \bU + \bX^\top\bX.
$
As a consequence, direct sampling from a DAG-Wishart posterior can be done using the \texttt{rDAGWishart} function simply by setting the input shape and rate parameters \texttt{a,U} as $\widetilde{a}$ and $\widetilde \bU$ respectively.

\subsection{DAG marginal likelihood}
\label{sec:marg:likelihood}

Finally, because of parameter prior independence in \eqref{eq:prior:indep:chol}, the marginal likelihood of DAG $\D$ admits the same node-by-node factorization of \eqref{eq:likelihood}, namely
\ben
\begin{aligned}
	m(\bX\g\D)
	&=
	\int f(\bX\g\bD,\bL,\D) \,p(\bD,\bL\g\D) \, d (\bD,\bL) \\
	&=
	\prod_{j=1}^{q}
	m(\bX_j\g\bX_{\pa_{\D}(j)},\D).
\end{aligned}
\een
In addition, because of conjugacy of the priors $p(\bD_{jj},\bL_{\prec j \, ]})$ with the
Normal densities $d\,\N_n(\bX_j\g\cdot)$ in \eqref{eq:likelihood}, each term $m(\bX_j\g\bX_{\pa_{\D}(j)},\D)$ can be obtained in closed-form expression from the ratio of prior and posterior normalizing constants as
\ben
\label{eq:marg:like:j}
m(\bX_j\g\bX_{\pa_{\D}(j)}, \D) =
(2\pi)^{-\frac{n}{2}}\cdot
\frac{\big|\bU_{\prec j \succ}\big|^{\frac{1}{2}}}
{\big|\widetilde{\bU}_{\prec j \succ}\big|^{\frac{1}{2}}}\cdot
\frac{\Gamma\left(\frac{1}{2}\widetilde{a}_j^{\D}\right)}
{\Gamma\left(\frac{1}{2}a_j^{\D}\right)}\cdot
\frac{\Big(\frac{1}{2}\bU_{jj\g\pa_{\D}(j)}\Big)^{\frac{1}{2}a_j^{\D}}}
{\left(\frac{1}{2}\widetilde{\bU}_{jj\g\pa_{\D}(j)}\right)^{\frac{1}{2}\widetilde{a}_j^{\D}}};
\een
Equation \eqref{eq:marg:like:j} is implemented in the internal function \texttt{DW\_nodelml(node, DAG, tXX, n, a, U)} which returns the logarithm of the node-marginal likelihood $m(\bX_j\g\bX_{\pa_{\D}(j)}, \D)$.
Arguments of the function are:
\begin{itemize}
	\item \texttt{node}: the node $j$;
	\item \texttt{DAG}: the adjacency matrix of the underlying DAG $\D$;
	\item \texttt{tXX}: the $(q,q)$ matrix $\bX^\top\bX$;
	\item \texttt{n}: the sample size $n$;
	\item \texttt{a,U}: the shape and rate hyperparameters of the compatible DAG-Wishart prior.
\end{itemize}

\subsection{Prior on DAGs}

To complete our Bayesian model specification, we finally assign a prior to each DAG $\D\in \mathcal{S}_q$, the set of all DAGs on $q$ nodes.
Specifically, we define a prior on $\D$ through independent Bernoulli distributions on the elements of the skeleton (underlying undirected graph) of $\D$. To this end, let $\bS^{\D}$ be the (symmetric) $0-1$ adjacency matrix of the skeleton of $\D$ whose $(u,v)$-element is denoted by $\bS^{\D}_{u,v}$.
For a given prior probability of edge inclusion $\pi\in[0,1]$,
we first assign a Bernoulli prior independently to each element $\bS^{\D}_{u,v}$ belonging to the lower-triangular part of $\bS^{\D}$, that is
$\bS^{\D}_{u,v} \, \stackrel{iid}{\sim} \, \textnormal{Ber}(\pi), u > v$.
As a consequence we obtain
\ben
p(\bS^{\D})=
\pi^{|\bS^{\D}|}(1-\pi)^{\frac{q(q-1)}{2}-|\bS^{\D}|},
\een
where $|\bS^{\D}|$ denotes the number of edges in the skeleton, or equivalently the number of entries equal to one in the lower-triangular part of $\bS^{\D}$.
Finally, we set
$p(\D)\propto p(\bS^{\D})$
for $\D \in \mathcal{S}_q$.





\section{Causal effects}
\label{sec:causal:effects}

In this section we summarize the definition of causal effect based on the \textit{do-calculus} theory \citep{Pear:2000}.
Let $I \subseteq V$ be an intervention \emph{target}.
A \emph{hard intervention}
on the set of variables $\{X_j, \,j\in I\}$
is denoted by
$\textnormal{do}\{X_j = \tilde{x}_j\}_{j\in I}$
and consists in the action of fixing each $X_j$, $j \in I$, to some chosen value $\tilde{x}_j$.
Graphically, the effect of an intervention on variables in $I$ is represented through the so-called \emph{intervention DAG} of $\D$, $\D^I$, which is obtained by removing all edges $(u,j)$ in $\D$ such that $j\in I$.
Under the Gaussian assumption \eqref{eq:Gaussian:DAG:Chol:facorization},
the consequent \emph{post-intervention} distribution can be written using the following \emph{truncated} factorization
\ben
\label{eq:post:intervention:gaussian}
f(\bx\g\textnormal{do}\{X_j = \tilde{x}_j\}_{j\in I}, \bD,\bL) = 
\begin{cases}
	\prod\limits_{i \notin I} d\,\N\big(x_i\g-\bL_{\prec i\,]}^{\top}\bx_{\pa_{\D}(i)}, \bD_{jj}\big)\big|_{\{x_j = \tilde{x}_j\}_{j\in I}}
	& \text{if }  x_j=\tilde{x}_j \,\, \forall j \in I, \quad
	\\
	\,\, 0 & \text{otherwise};
\end{cases}
\een
see also \citet{Pear:2000} and \citet{Maat:etal:2009} for details.
For a given ``response" variable $Y \in \{X_1,\dots,X_q\}$, \citet{nandy2017estimating} define the \emph{total} joint effect of an intervention $\text{do}\{X_j = \tilde{x}_j\}_{j \in I}$ on $Y$ as
\begin{equation}
	\label{jointcausaleffect}
	\bm{\theta}_{Y}^{I} := \big(\theta_{h,Y}^{I}\big)_{h \in I},
\end{equation}
where, for each $h \in I$,
\begin{equation}
	\theta_{h,Y}^I := \frac{\partial}{\partial{x_h}} \vat\big(Y\g\text{do}\{X_j = \tilde{x}_j\}_{j \in I}\big)
\end{equation}
is the causal effect on $Y$ associated to variable $X_h$ in the joint intervention.
In the Gaussian setting of Equation \eqref{eq:post:intervention:gaussian} we obtain
\ben
\label{eq:post:intervention:gaussian:bis}
X_1,\dots,X_q\g \textnormal{do}\{X_j = \tilde{x}_j\}_{j\in I},\bOmega
\, \sim \, \N_q\big(\bzero, (\bOmega^{I})^{-1}\big),
\een
where
$
\bOmega^{I}=(\bL^I)\bD^{-1}(\bL^I)^{\top}
$
and
\ben
\label{eq:from:L:to:L:int}
\bL^I_{u,v} = 
\begin{cases}
	\,\, 0  & \text{if } v \in I \text{ and } v\ne u
	\\
	\bL_{u,v} & \text{otherwise}.
\end{cases}
\een
Finally, the causal effect of $X_h$ on $Y$ $(h\in I)$ in a joint intervention on $\{X_j\}_{j \in I}$ is given by
\ben
\label{eq:causal:effect:gaussian}
\theta_{h,Y}^I = \bSigma^I_{h,Y}\big(\bSigma^I_{h,h}\big)^{-1},
\een
where $\bSigma^I=(\bOmega^I)^{-1}$; see also \citet{nandy2017estimating}.
It follows that the causal effect $\theta_{h,Y}^I$ is a function of the precision matrix $\bOmega$ (equivalently, $\bL$ and $\bD$) which in turn depends on the underlying DAG $\D$.

In our package, the total causal effect on $Y$ of an intervention on variables $\{X_j\}_{j\in I}$
can be computed using the function \texttt{causaleffect(targets, response, L, D)} whose arguments are:
\begin{itemize}
	\item \texttt{targets}: a vector with the numerical labels of the intervened nodes $I$;
	\item \texttt{response}: the numerical label of the response variable of interest $Y$;
	\item \texttt{L, D}: the DAG-parameters $\bL$ and $\bD$.
\end{itemize}

Moving back to the example in Section \ref{subsec:DAG-Wishart:sampling}, we can compute the total causal effect on $Y=X_1$ of an intervention with target $I=\{3,4\}$ as

\begin{verbatim}
	causaleffect(targets = c(3,4), response = 1, L = outDL$L, D = outDL$D)
	[1]  1.65984864 -0.06790017
\end{verbatim}

The two coefficients correspond to the causal effects $\theta_{3,1}^I, \theta_{4,1}^I$ computed as in Equation \eqref{eq:causal:effect:gaussian} for $I=\{3,4\}$, $Y=X_1$.

\section{MCMC scheme}
\label{sec:MCMC}

In this section we briefly summarize the MCMC scheme that we adopt to target the joint posterior distribution 
of DAG structures and DAG-parameters,
\ben
\label{eq:posterior}
p(\bD, \bL,\D \g \bX)
\propto
f_{}(\bX \g \bD, \bL, \D)\,p(\bD, \bL \g \D)\,p(\D),
\een
where $\bX$ is the $(n,q)$ data matrix.
The proposed sampler is based on a reversible jump MCMC algorithm which takes into account the partial analytic structure
(PAS, \citealt{Godsill:2012}) of the DAG-Wishart distribution to sample DAG $\D$ and DAG-parameters $(\bD,\bL)$
from their full conditional distributions. For further details the reader can refer to \citet[Supplementary Material]{Caste:Consonni:2021:BA}.

\subsection{Proposal over the DAG space}
\label{sec:proposal:DAGs}

\begin{figure}
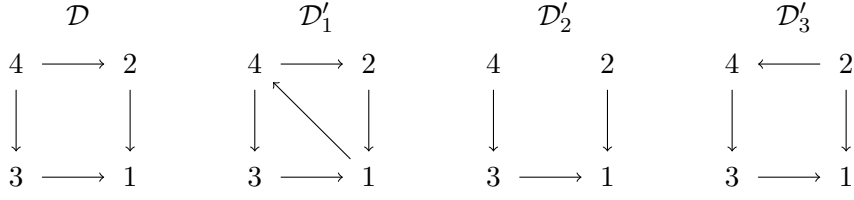

	\begin{center}
		\begin{tabular}{ccccccc}
			$\D$ && $\D'_1$ && $\D'_2$ && $\D'_3$\\
			{\normalsize
				\tikz \graph [no placement, math nodes, nodes={circle}]
				{
					4[x=0,y=0] -> {2[x=1.5,y=0], 3[x=0,y=-1.5]},
					{2[x=1.5,y=0], 3[x=0,y=-1.5]} -> 1[x=1.5,y=-1.5]};
			} & &
			{\normalsize
				\tikz \graph [no placement, math nodes, nodes={circle}]
				{
					4[x=0,y=0] -> 3[x=0,y=-1.5],
					4[x=0,y=0] -> 2[x=1.5,y=0], 
					4[x=0,y=0] <- 1[x=1.5,y=-1.5],
					{2[x=1.5,y=0], 3[x=0,y=-1.5]} -> 1[x=1.5,y=-1.5]};
			} & &
			{\normalsize
				\tikz \graph [no placement, math nodes, nodes={circle}]
				{
					4[x=0,y=0] -> {3[x=0,y=-1.5]}, 
					{2[x=1.5,y=0], 3[x=0,y=-1.5]} -> 1[x=1.5,y=-1.5]};
			} & &
			{\normalsize
				\tikz \graph [no placement, math nodes, nodes={circle}]
				{
					4[x=0,y=0] <- 2[x=1.5,y=0], 
					4[x=0,y=0] -> 3[x=0,y=-1.5],
					{2[x=1.5,y=0], 3[x=0,y=-1.5]} -> 1[x=1.5,y=-1.5]};
			}
		\end{tabular}
	\end{center}
	\caption{A DAG $\D$ and three modified graphs of the operators InsertD $1\rightarrow 4$, DeleteD $4\rightarrow 2$, ReverseD $4\rightarrow 2$ respectively. Operator InsertD $1\rightarrow 4$ is \textit{not} valid since $\D'_1$ is not acyclic.}
	\label{fig:DAGs}
\end{figure}

First step of our algorithm is the definition of a proposal distribution $q(\D'\g\D)$ determining the transitions between DAGs within the space $\mathcal{S}_q$.
To this end we consider three types of operators that locally modify an input DAG $\D$: insert a directed edge (InsertD $u\rightarrow v$ for short), delete a directed edge (DeleteD $u\rightarrow v$) and reverse a directed edge (ReverseD $u\rightarrow v$); see Figure \ref{fig:DAGs} for an example.

For any $\D\in\mathcal{S}_q$, we can construct the set of \textit{valid} operators $\mathcal{O}_{\D}$, that is operators whose resulting graph \emph{is} a DAG.
Given a current DAG $\D$ we then propose $\D'$ by uniformly sampling a DAG in $\mathcal{O}_\D$.
The construction of $\mathcal{O}_{\D}$ and the DAG proposal are summarized in Algorithm \eqref{alg:propose:DAG}.
Also notice that because there is a one-to-one correspondence between each operator and resulting DAG $\D'$,
the probability of transition from $\D$ to $\D'$ (a \textit{direct successor} of $\D$) is
$q(\D'\g\D) = 1/|\mathcal{O}_{\D}|$.

\begin{algorithm}{
		\SetAlgoLined
		\vspace{0.1cm}
		\KwInput{A DAG $\D$}
		\KwOutput{The set of valid operators $\mathcal{O}_{\D}$, a DAG $\D'$ drawn from $q(\D'\g\D)$}
		Construct:\\
		$I_{\D}$, the set of all possible operators of type InsertD,\\
		$E_{\D}$, the set of all possible operators of type DeleteD,\\ $R_{\D}$, the set of all possible operators of type ReverseD\;
		\For{each operator $o_D\in I_{\D}$}{
			add $o_{\D}$ to $\mathcal{O}_{\D}$ if $o_{\D}$ is valid}
		\For{each operator $o_D\in E_{\D}$}{
			add $o_{\D}$ to $\mathcal{O}_{\D}$ if $o_{\D}$ is valid}
		\For{each operator $o_D\in R_{\D}$}{
			add $o_{\D}$ to $\mathcal{O}_{\D}$ if $o_{\D}$ is valid}
		Draw uniformly an operator $o_{\D}$ from $\mathcal{O}_{\D}$ and obtain $\D'$ by applying it to $\D$
	}
	\caption{Construction of $\mathcal{O}_{\D}$ and sampling of $\D'$ from $q(\D'\g\D)$}
	\label{alg:propose:DAG}
\end{algorithm}

\subsection{DAG update}
\label{sec:MCMC:update:DAG}

Because of the structure of the proposal distribution introduced in Section \ref{sec:proposal:DAGs}, at each step of our MCMC algorithm we will need to compare two DAGs $\D$ and $\D'$ which differ by one edge only. Notice that operator ReverseD $u\rightarrow v$ can be also brought back to the same case since is equivalent to the consecutive application of the operators DeleteD $u\rightarrow v$ and InsertD $v\rightarrow u$.
Therefore, consider two DAGs $\D=(V,E)$, $\D'=(V,E')$ such that $E'=E\setminus\{(h,j)\}$.
We also index each parameter with its own DAG-model and write accordingly
$(\bD^{\D}, \bL^{\D})$ and $(\bD^{\D'}, \bL^{\D'})$.
The two sets of parameters under the DAGs $\D,\D'$ differ only with regard to their $j$-th component
$(\bD_{jj}^{\D}, \bL^{\D}_{\prec j\,]})$, and $(\bD_{jj}^{\D'}, \bL^{\D'}_{\prec j\,]})$ respectively.
Moreover the remaining  parameters
$\{ \bD_{rr}^{\D}, \bL^{\D}_{\prec r\,]}; \, r \neq j \}$ and
$\{ \bD_{rr}^{\D'}, \bL^{\D'}_{\prec r\,]}; \, r \neq j \}$
are componentwise \emph{equivalent} between the two graphs because they refer to structurally equivalent  conditional models; see in particular Equation
\eqref{eq:Gaussian:DAG:Chol:facorization}.
This is crucial for the correct application of the PAS algorithm; see also \citet{Godsill:2012}.
The acceptance probability for $\D'$ under a PAS algorithm is then given by
$\alpha_{\D'}=\min\{1;r_{\D'}\}$,
where
\ben
\label{eq:ratio:PAS:DAG}
r_{\D'}
&=&\frac{p(\D'\g\bD^{\D'}\setminus \bD_{jj}^{\D'},\bL^{\D'}\setminus \bL^{\D'}_{\prec j\,]},\bX)}
{p(\D\g\bD^{\D}\setminus \bD_{jj}^{\D},\bL^{\D}\setminus \bL^{\D}_{\prec j\,]},\bX)}
\cdot
\frac{q(\D\g\D')}{q(\D'\g\D)}
\nonumber \\
&=&
\frac{p(\bX,\bD^{\D'}\setminus \bD_{jj}^{\D'},\bL^{\D'}\setminus \bL^{\D'}_{\prec j\,]}\g \D')}
{p(\bX,\bD^{\D}\setminus \bD_{jj}^{\D},\bL^{\D}\setminus \bL^{\D}_{\prec j\,]}\g \D)}
\cdot\frac{p(\D')}{p(\D)}
\cdot\frac{q(\D\g\D')}{q(\D'\g\D)}.
\een
Therefore we require to evaluate for DAG $\D$
\be
p(\bX,\bD\setminus \bD_{jj},\bL\setminus \bL_{\prec j\,]}\g \D)
&=&
\int_0^{\infty}\int_{\Re^{|\pa_{\D}(j)|}}
p(\bX\g\bD,\bL,\D)p(\bD,\bL\g\D) \, d \bL_{\prec j\,]} d\bD_{jj};
\ee
similarly for $\D'$.
Moreover, because of the likelihood and prior factorizations in \eqref{eq:likelihood} and \eqref{eq:prior:indep:chol} we can write
\ben
\label{eq:PAS:joint:DAG}
p(\bX,\bD\setminus \bD_{jj},\bL\setminus \bL_{\prec j\,]}\g \D)
&=&
\prod_{r\ne j}f(\bX_r\g\bX_{\pa_{\D}(r)},\bD_{rr},\bL_{\prec r\,]},\D)p(\bL_{\prec r\,]}\g\bD_{rr},\D)p(\bD_{rr}\g\D) \nonumber \\
&\cdot&
\int_0^{\infty}\int_{\Re^{|\pa_{\D}(j)|}}
f(\bX_j\g\bX_{\pa_{\D}(j)},\bD_{jj},\bL_{\prec j\,]},\D) \\
&\cdot&
p(\bL_{\prec j\,]}\g\bD_{jj},\D)p(\bD_{jj}\g\D)
\, d \bL_{\prec j\,]} d\bD_{jj}. \nonumber
\een
Finally, the integral in \eqref{eq:PAS:joint:DAG} corresponds to the node-marginal likelihood $m(\bX_j\g\bX_{\pa_{\D}(j)},\D)$ which is available in the closed form \eqref{eq:marg:like:j}. 
Therefore, the acceptance rate \eqref{eq:ratio:PAS:DAG} simplifies to
\ben
\label{eq:ratio:PAS:DAG:simplified}
r_{\D'} =
\frac{m_{}(\bX_j\g\bX_{\pa_{\D'}(j)}, \D')}
{m_{}(\bX_j\g\bX_{\pa_{\D}(j)}, \D)}
\cdot\frac{p(\D')}{p(\D)}
\cdot\frac{q(\D\g\D')}{q(\D'\g\D)}.
\een

\begin{figure}
	\begin{center}
		\includegraphics[height=4.4cm,width=5.1cm]{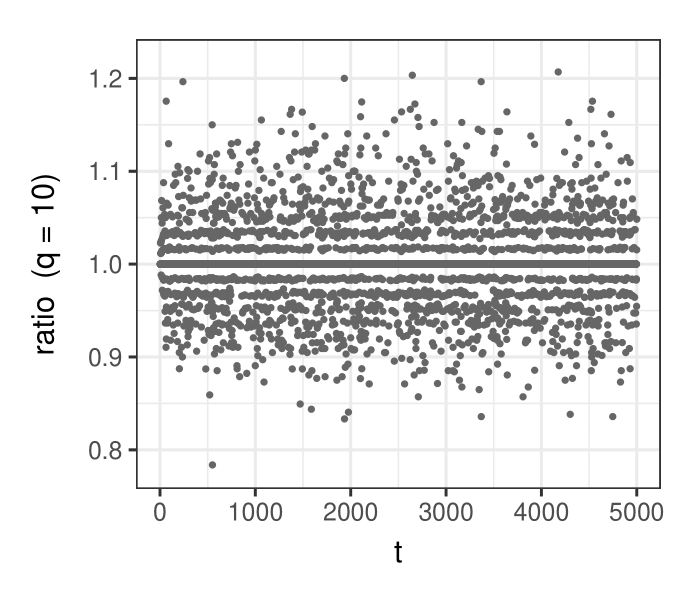}
		\includegraphics[height=4.4cm,width=5.1cm]{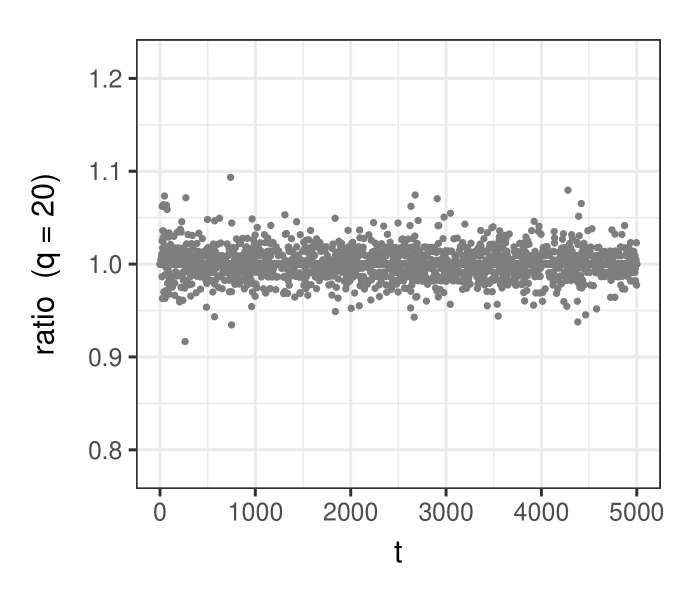}
		\includegraphics[height=4.4cm,width=5.1cm]{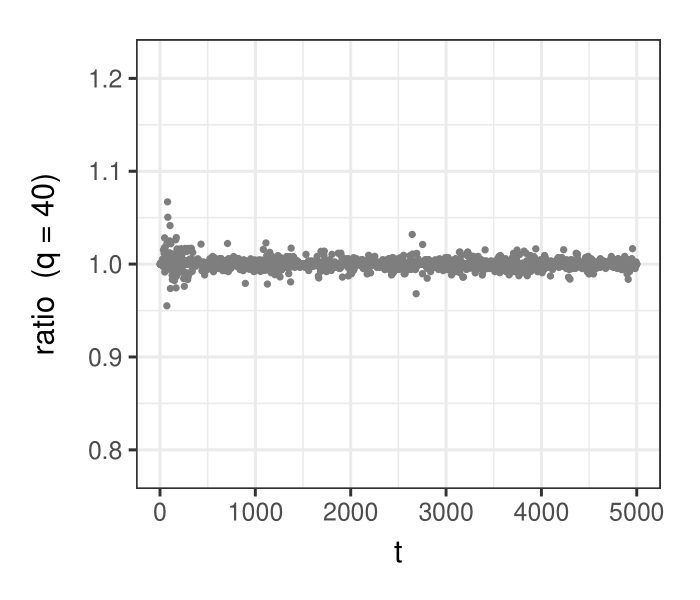}
		\caption{\small Ratio of proposal distributions $q(\D\g\D')/q(\D'\g\D)$ for a Markov chain on the DAG space $\mathcal{S}_q$ generated using Algorithm \ref{alg:propose:DAG}. Each dot is the ratio of proposals for the current DAG $\D$ and its sampled successor $\D'$. Left to right-side panels refer to $q\in\{10,20,40\}$ respectively.}
		\label{fig:ratio:approx}
	\end{center}
\end{figure}

Notice that to compute \eqref{eq:ratio:PAS:DAG:simplified} we need to evaluate the ratio of proposals $q(\D\g\D')/q(\D'\g\D)=|\mathcal{O}_{\D}|/|\mathcal{O}_{\D'}|$ which in turn requires the construction of the sets of valid operators $\mathcal{O}_{\D}, \mathcal{O}_{\D'}$. This step can be computationally expensive especially when $q$ is large. However, we show in the following simulation that for adjacent DAGs, namely DAGs differing by the insertion/deletion/reversal of one edge only, the approximation $q(\D\g\D')/q(\D'\g\D)=1$ is reasonable and also becomes as accurate as $q$ increases.
To this end, we vary the number of nodes $q\in\{10,20,40\}$. For each value of $q$, starting from the empty (null) DAG $\D$, we first construct $\mathcal{O}_{\D}$ and randomly sample an operator $o_{\D}\in\mathcal{O}_{\D}$ as in Algorithm \ref{alg:propose:DAG}; next, we apply $o_{\D}$ to $\D$ and obtain $\D'$; we finally construct $\mathcal{O}_{\D'}$ and compute the ratio $|\mathcal{O}_{\D}|/|\mathcal{O}_{\D'}|$. By reiterating this procedure $T = 5000$ times we get a Markov chain over the space of DAGs $\mathcal{S}_q$ and $T$ values of the ratio. Results for each number of nodes $q$ are summarized in the scatter-plots  of Figure \ref{fig:ratio:approx}. By inspection it is clear that the proposed approximation becomes as precise as $q$ grows since points are more and more concentrated around the one value.

By adopting the proposed approximation, instead of testing the validity of each (possible) operator in $I_{\D},E_{\D},R_{\D}$ (needed to build $\mathcal{O}_{\D}$), we can simply draw one operator among the set of possible operators $\{I_{\D} \cup E_{\D} \cup R_{\D}\}$ and apply it to the input DAG $\D$ whenever valid. This simplification leads to Algorithm \ref{alg:propose:DAG:fast}, the fast version of Algorithm \ref{alg:propose:DAG}.
We finally recommend this approximation for moderate-to-large number of nodes.

\begin{algorithm}{
		\SetAlgoLined
		\vspace{0.1cm}
		\KwInput{A DAG $\D$}
		\KwOutput{A DAG $\D'$ approximately drawn from $q(\D'\g\D)$}
		Construct:\\
		$I_{\D}$, the set of all possible operators of type InsertD,\\
		$E_{\D}$, the set of all possible operators of type DeleteD,\\ $R_{\D}$, the set of all possible operators of type ReverseD\;
		Draw uniformly an operator $o_{\D}\in \{I_{\D}\cup E_{\D}\cup R_{\D}\} \cup R_{\D}$\;
		\While{$o_{\D}$ is not valid}{
			Draw uniformly a new $o_{\D}\in \{I_{\D}\cup E_{\D} \cup R_{\D}$\}}
		Apply $o_{\D}$ to $\D$ and obtain $\D'$
	}
	\caption{Approximate sampling of $\D'$ from $q(\D'\g\D)$}
	\label{alg:propose:DAG:fast}
\end{algorithm}

\subsection{Update of DAG-parameters}

In the second step of the MCMC scheme we then sample the model-dependent parameters $(\bD,\bL)$
conditionally on the accepted DAG $\D$ from their full conditional distribution
\ben
\label{eq:posterior:update}
p(\bD,\bL\g\D,\bX) = 
\prod_{j=1}^q
p(\bL_{\prec j\,]}\g\bD_{jj}, \D, \bX)\,
p(\bD_{jj} \g \D, \bX).
\een
Equation \eqref{eq:posterior:update} corresponds to the DAG-Wishart posterior distribution in \eqref{eq:posterior:chol}.
As mentioned, direct sampling from $p(\bD,\bL\g\D,\bX)$ can be performed using the \texttt{rDAGWishart} function introduced in Section \ref{subsec:DAG-Wishart:sampling}; see also Section \ref{subsec:DAG-Wishart:posterior}.

\subsection{MCMC algorithm}

Our main MCMC scheme for posterior inference on DAGs and DAG-parameters is summarized in Algorithm \ref{alg:MCMC}.

\begin{algorithm}{
		\SetAlgoLined
		\vspace{0.1cm}
		\KwInput{$S$, $B$, $\bX$, $a,\bU, w$}
		\KwOutput{$S$ samples from the posterior distribution \eqref{eq:posterior}}
		Initialize $\D^{(0)}$, e.g. the empty DAG\;
		\For{$s=1,\dots,B+S$}{
			Sample $\D'$ from $q(\D'\g\D^{(s-1)})$ using Algorithm \ref{alg:propose:DAG} or its fast version (Algorithm \ref{alg:propose:DAG:fast})\;
			Set $\D^{(s)}=\D'$ with probability $\alpha_{\D'}=\min\{1;r_{\D'}\}$ in \eqref{eq:ratio:PAS:DAG:simplified},
			otherwise $\D^{(s)}=\D^{(s-1)}$\;
			Sample $\big(\bD^{(s)}, \bL^{(s)}\big)$ from its full conditional \eqref{eq:posterior:update}\;
		}
		\Return $\big\{\big(\bD^{(s)},\bL^{(s)},\D^{(s)} \big), s=B+1,\dots,B+S\big\}$
	}
	\caption{MCMC scheme to sample from the posterior of DAGs and parameter}
	\label{alg:MCMC}
\end{algorithm}

Algorithm \ref{alg:MCMC} is implemented within our package in the function
\texttt{learn\_DAG(S, burn, data, a, U, w, fast = FALSE, save.memory = FALSE, collapse = FALSE)}. Main arguments of the function are:
\begin{itemize}
	\item \texttt{S}: the number of final MCMC draws from the posterior;
	\item \texttt{burn}: the burn-in period;
	\item \texttt{data}: the $(n,q)$ data matrix $\bX$;
	\item \texttt{a, U}: the hyperparameters of the DAG-Wishart prior;
	\item \texttt{w}: the prior probability of edge inclusion.
\end{itemize}

The default setting \texttt{fast = FALSE} implements the MCMC proposal distribution of Algorithm \ref{alg:propose:DAG} which requires the enumeration of all direct successor DAGs in each local move. By converse, with \texttt{fast = TRUE} the approximate proposal of Algorithm \ref{alg:propose:DAG:fast} is adopted.

\vspace{0.2cm}

Whenever the interest is on DAG structure learning only, rather than on both DAG learning and parameter inference, a \emph{collapsed} MCMC sampler can be considered.
In such a case, the target distribution is represented by the \textit{marginal} posterior over the DAG space
\ben
\label{eq:posterior:collapsed}
p(\D \g \bX)
\propto
m_{}(\bX \g \D)\,p(\D),
\een
where $m_{}(\bX \g \D)$ is the marginal likelihood of DAG $\D$ derived in Section \ref{sec:marg:likelihood}.
An MCMC sampler targeting the posterior $p(\D \g \bX)$, $\D \in \mathcal{S}_q$, can be constructed as in Algorithm 
\ref{alg:MCMC:collapsed} and implemented by \texttt{learn\_DAG} by setting \texttt{collapse = TRUE}.

\begin{algorithm}{
		\SetAlgoLined
		\vspace{0.1cm}
		\KwInput{$S$, $B$, $\bX$, $a,\bU, w$}
		\KwOutput{$S$ samples from the posterior in \eqref{eq:posterior}}
		Initialize $\D^{(0)}$, e.g. the empty DAG\;
		\For{$s=1,\dots,B+S$}{
			Sample $\D'$ from $q(\D'\g\D^{(s-1)})$ using Algorithm \ref{alg:propose:DAG} or its fast version (Algorithm \ref{alg:propose:DAG:fast})\;
			Set $\D^{(s)}=\D'$ with probability $\alpha_{\D'}=\min\{1;r_{\D'}\}$ in \eqref{eq:ratio:PAS:DAG:simplified},
			otherwise $\D^{(s)}=\D^{(s-1)}$\;
		}
		\Return $\big\{\D^{(s)}, s=B+1,\dots,B+S\big\}$
	}
	\caption{Collapsed MCMC scheme to sample from the posterior of DAGs}
	\label{alg:MCMC:collapsed}
\end{algorithm}

By default, with \texttt{save.memory = FALSE}, \texttt{learn\_DAG} returns a list of $(q,q,S)$-dimensional arrays. However, as the dimension of the chain increases, the user may incur in memory usage restrictions. By setting \texttt{save.memory = TRUE}, at each iteration the sampled matrices are converted into character strings and stored in a vector, thus reducing the size of the output. Function \texttt{bd\_decode} also allows to convert the output into the original array format.


\subsection{Posterior inference}
\label{sec:posterior:inference}

Algorithm \ref{alg:MCMC} produces a collection of DAGs $\big\{\D^{(s)}\big\}_{s=1}^{S}$ and DAG-parameters $\big\{\big(\bD^{(s)},\bL^{(s)}\big)\big\}_{s=1}^{S}$
visited by the MCMC algorithm which can be used to approximate the target distribution in \eqref{eq:posterior}.
An approximate posterior distribution over the space $\mathcal{S}_q$ can be obtained as
\ben
\label{eq:posterior:dags}
\hat{p}(\D\g\bX)=\frac{1}{S}\sum_{s=1}^{S}\mathbbm{1}\big\{\D^{(s)}=\D\big\},
\een
that is through the MCMC frequency of visits of each DAG $\D$.
In addition, we can compute, for each $(u,v)$, $u \ne v$, the posterior probability of edge inclusion
\ben
\label{eq:posterior:edge:inclusion}
\hat p_{u\rightarrow v}(\bX)\equiv \hat p_{u\rightarrow v}=\frac{1}{S}\sum_{s=1}^{S}\mathbbm{1}_{u\rightarrow v}\big\{\D^{(s)}\big\},
\een
where $\mathbbm{1}_{u\rightarrow v}\big\{\D^{(s)}\big\}$ takes value 1 if and only if $\D^{(s)}$ contains the edge $u\rightarrow v$.
A $(q,q)$ matrix collecting the posterior probabilities of edge inclusion can be constructed through the function
\texttt{get\_edgeprobs(learnDAG\_output)} which requires as the only argument the output of \texttt{learn\_DAG}.

Single DAG estimates summarizing the MCMC output can be also recovered.
For instance, one can consider the Maximum A Posteriori (MAP) DAG estimate, corresponding to the DAG with the highest estimated posterior probability. Alternatively, we can construct the Median Probability (DAG) Model (MPM) which is instead obtained by including all edges $u\rightarrow v$ whose estimated probability of inclusion \eqref{eq:posterior:edge:inclusion} exceeds 0.5.
The two DAG estimates can be constructed through the functions \texttt{get\_MAPdag(learnDAG\_output)} and
\texttt{get\_MPMdag(learnDAG\_output)} respectively.

\vspace{0.2cm}

Given the MCMC output, we can recover the collection of precision matrices $\big\{\bOmega^{\D^{(s)}}\big\}_{s=1}^{S}$ using the relationship $\bOmega=\bL\bD^{-1}\bL^\top$.
For a given intervention target $I \subseteq \{1,\dots,q\}$, recall now the definition of causal effect of $X_h$ on $Y$ $(h\in I)$ in a joint intervention on $\{X_j\}_{j \in I}$ summarized in Section \ref{sec:causal:effects}.
Draws from the posterior distribution of each causal effect coefficient $\theta_{h,Y}^I$, $h \in I$, can be then recovered using \eqref{eq:causal:effect:gaussian}. This is performed by function \texttt{get\_causaleffect(learnDAG\_output, targets, response, BMA = FALSE)}
under the default choice \texttt{BMA = FALSE}. Arguments of the function are:
\begin{itemize}
	\item \texttt{learnDAG\_output}: the output of \texttt{learn\_dag};
	\item \texttt{targets}: a vector with the numerical labels of the intervened nodes in $I$;
	\item \texttt{response}: the numerical label of the response variable of interest $Y$.
\end{itemize}
Its output therefore consists of $S$ draws $\big\{\theta_{h,Y}^{I(1)}, \dots, \theta_{h,Y}^{I(S)}\big\}$ approximately sampled from the posterior of each causal effect coefficient $\theta_{h,Y}^I$, $h \in I$.
As a summary of the posterior distribution of each $\theta_{h,Y}^I$ we can further consider
\ben
\label{eq:BMA:causal:effect}
\hat{\theta}_{h,Y}^{I} =
\frac{1}{S} \sum_{s=1}^{S}
\theta_{h,Y}^{I(s)},
\een
which corresponds to a Bayesian Model Averaging (BMA) estimate wherein posterior model probabilities are approximated through their MCMC frequencies of visits.
BMA estimates of causal effect coefficients are returned by \texttt{get\_causaleffects} when \texttt{BMA = TRUE}.

\subsection{MCMC diagnostics of convergence}
\label{sec:MCMC:diagnostics}

MCMC diagnostics of convergence can be performed by monitoring how specific graph features vary across MCMC iterations. Selected diagnostics that we now detail are implemented by function \texttt{get\_diagnostics}, which takes as input the MCMC output of \texttt{learn\_DAG}.

We start by focusing on the number of edges in the DAG.
The first ouput of \texttt{get\_diagnostics} is a trace plot of the number of edges in the DAGs visited by the MCMC chain at each step $s = 1, ..., S$.
The absence of trends in the plot generally suggests a good degree of MCMC mixing.
In addition, a trace plot of the average number of edges in the DAGs visited up to time $s$, for $s = 1, ..., S$, is returned. In this case,
the convergence of the plot around a ``stable" average size represents a symptom of genuine MCMC convergence.

We further monitor the posterior probability of edge inclusion computed, for each edge $u \rightarrow v$, up to time $s$, for $s = 1, \dots, S$.
Each posterior probability is estimated as the proportion of DAGs visited by the MCMC up to time $s$ which contain the directed edge $u \rightarrow v$;
see also Equation \eqref{eq:posterior:edge:inclusion}.
Output is organized in $q$ plots (one for each node $v = 1, \dots, q$), each summarizing the posterior probabilities of edges $u \rightarrow v$, $u = 1, \dots, q$.
The stabilization of each posterior probability around a constant level reflects a good degree of MCMC convergence; see also the following section \ref{sec:illustrations} for more detailed illustrations.

\section{Illustrations} \label{sec:illustrations}

In this section we exemplify the use of \textbf{\texttt{BCDAG}} through both simulated and real data.

\subsection{Simulated data}

We first use the function \texttt{rDAG(q, w)} to randomly generate a DAG structure with $q=8$ nodes by fixing a probability of edge inclusion $w = 0.2$. 

\begin{verbatim}
	q <- 8
	set.seed(123)
	DAG <- rDAG(q = q, w = 0.2)
	DAG
	1 2 3 4 5 6 7 8
	1 0 0 0 0 0 0 0 0
	2 0 0 0 0 0 0 0 0
	3 0 1 0 0 0 0 0 0
	4 0 0 0 0 0 0 0 0
	5 1 0 0 0 0 0 0 0
	6 1 1 1 1 0 0 0 0
	7 0 0 0 1 1 0 0 0
	8 0 0 0 0 0 0 0 0
\end{verbatim}

Given the so-obtained DAG, above represented through its adjacency matrix \texttt{DAG}, we then generate the non-zero off-diagonal elements of $\bL$ uniformly in the interval $[0.1,1]$, while we fix $\bD = \bI_q$.

\begin{verbatim}
	L <- matrix(runif(n = q*(q-1), min = 0.1, max = 1), q, q)*DAG; diag(L) <- 1
	D <- diag(1, q)
\end{verbatim}

Finally, we generate $n = 200$ multivariate zero-mean Gaussian data as in \eqref{eq:Gaussian:DAG:model}, by recovering first the precision matrix $\bOmega=\bL\bD^{-1}\bL^{\top}$ and then using the function \texttt{rmvnorm} within the package \texttt{mvtnorm}:

\begin{verbatim}
	Omega <- L%*%solve(D)%*%t(L)
	X <- mvtnorm::rmvnorm(n = 200, sigma = solve(Omega))
\end{verbatim}

We run function \texttt{learn\_DAG} to approximate the posterior over DAGs and DAG-parameters (\texttt{collapse = FALSE}) by fixing the number of final MCMC iterations and burn-in period as $S=5000$, $B=1000$, while prior hyperparameters as $a = q, \bU=\bI_q, w = 0.1$.
We implement the approximate MCMC proposal (Section \ref{sec:MCMC:update:DAG}) by setting \texttt{fast = TRUE}:

\begin{verbatim}
	out_mcmc <- learn_DAG(S = 5000, burn = 1000, data = X, a = q, U = diag(1,q),
	            w = 0.1, fast = TRUE, save.memory = FALSE, collapse = FALSE)
\end{verbatim}

\texttt{out\_mcmc}, the output of function \texttt{learn\_DAG}, corresponds to a list with three elements:
\begin{itemize}
	\item \texttt{out\_mcmc\$graphs}: a $(q,q,S)$ array with the adjacency matrices of DAGs $\D^{(1)}, \dots,$ $\D^{(S)}$ sampled by the MCMC;
	\item \texttt{out\_mcmc\$L}: a $(q,q,S)$ array with the $S$ matrices $\bL^{(1)}, \dots, \bL^{(S)}$ sampled by the MCMC;
	\item \texttt{out\_mcmc\$D}: a $(q,q,S)$ array with the $S$ matrices $\bD^{(1)}, \dots, \bD^{(S)}$ sampled by the MCMC.
\end{itemize}

Given the MCMC output, we can compute the posterior probabilities of edge inclusion using function \texttt{get\_edgeprobs}. The latter returns a $(q,q)$ matrix with $(u,v)$-element corresponding to the estimated posterior probability $\hat p_{u\rightarrow v}$ in Equation \eqref{eq:posterior:edge:inclusion}. In addition, the MAP and MPM DAG estimates can be recovered through \texttt{get\_MAPdag} and \texttt{get\_MPMdag} respectively:

\begin{verbatim}
	get_edgeprobs(learnDAG_output = out_mcmc)
	
	
	       1      2      3      4      5      6      7      8
	1 0.0000 0.0000 0.0072 0.0012 0.0000 0.0000 0.0066 0.0038
	2 0.0000 0.0000 0.1084 0.0084 0.0352 0.9182 0.0000 0.0258
	3 0.0280 0.0504 0.0000 0.0828 0.0000 0.8872 0.0068 0.1822
	4 0.0008 0.0000 0.0126 0.0000 0.0136 0.0174 0.0292 0.0158
	5 1.0000 0.0082 0.0000 0.0028 0.0000 0.0000 0.4344 0.1360
	6 1.0000 0.0558 0.1128 0.9826 0.0000 0.0000 0.0330 0.0110
	7 0.0006 0.0044 0.0326 0.9708 0.5656 0.0214 0.0000 0.0360
	8 0.0154 0.0206 0.0666 0.0142 0.1056 0.0036 0.0136 0.0000
\end{verbatim}

\begin{verbatim}
	get_MAPdag(learnDAG_output = out_mcmc)
	
	  1 2 3 4 5 6 7 8
	1 0 0 0 0 0 0 0 0
	2 0 0 0 0 0 1 0 0
	3 0 0 0 0 0 1 0 0
	4 0 0 0 0 0 0 0 0
	5 1 0 0 0 0 0 0 0
	6 1 0 0 1 0 0 0 0
	7 0 0 0 1 1 0 0 0
	8 0 0 0 0 0 0 0 0
\end{verbatim}

\begin{verbatim}	
	get_MPMdag(learnDAG_output = out_mcmc)
	
	  1 2 3 4 5 6 7 8
	1 0 0 0 0 0 0 0 0
	2 0 0 0 0 0 1 0 0
	3 0 0 0 0 0 1 0 0
	4 0 0 0 0 0 0 0 0
	5 1 0 0 0 0 0 0 0
	6 1 0 0 1 0 0 0 0
	7 0 0 0 1 1 0 0 0
	8 0 0 0 0 0 0 0 0
\end{verbatim}

Suppose now we are interested in evaluating the total causal effect on node $Y=X_1$ consequent to a joint intervention on variables in $I = \{5,6,7\}$.
An approximate posterior distribution for the vector parameter $\big(\theta^I_{h,Y}\big)_{h\in I}$ representing the total causal effect (Section \ref{sec:causal:effects}) can be recovered from the MCMC output as

\begin{verbatim}
	joint_causal <- get_causaleffect(learnDAG_output = out_mcmc,
	                targets = c(5,6,7), response = 1, BMA = FALSE)
	
	head(joint_causal)
	
	          h = 5      h = 6 h = 7
	[1,] -0.6323857 -0.7955013     0
	[2,] -0.5649460 -0.8802489     0
	[3,] -0.5961280 -0.8219784     0
	[4,] -0.5790337 -0.8392289     0
	[5,] -0.6657683 -0.8439536     0
	[6,] -0.6334429 -0.9030620     0
\end{verbatim}

\begin{figure}[h]
	\centering
	\includegraphics[width=0.6\linewidth]{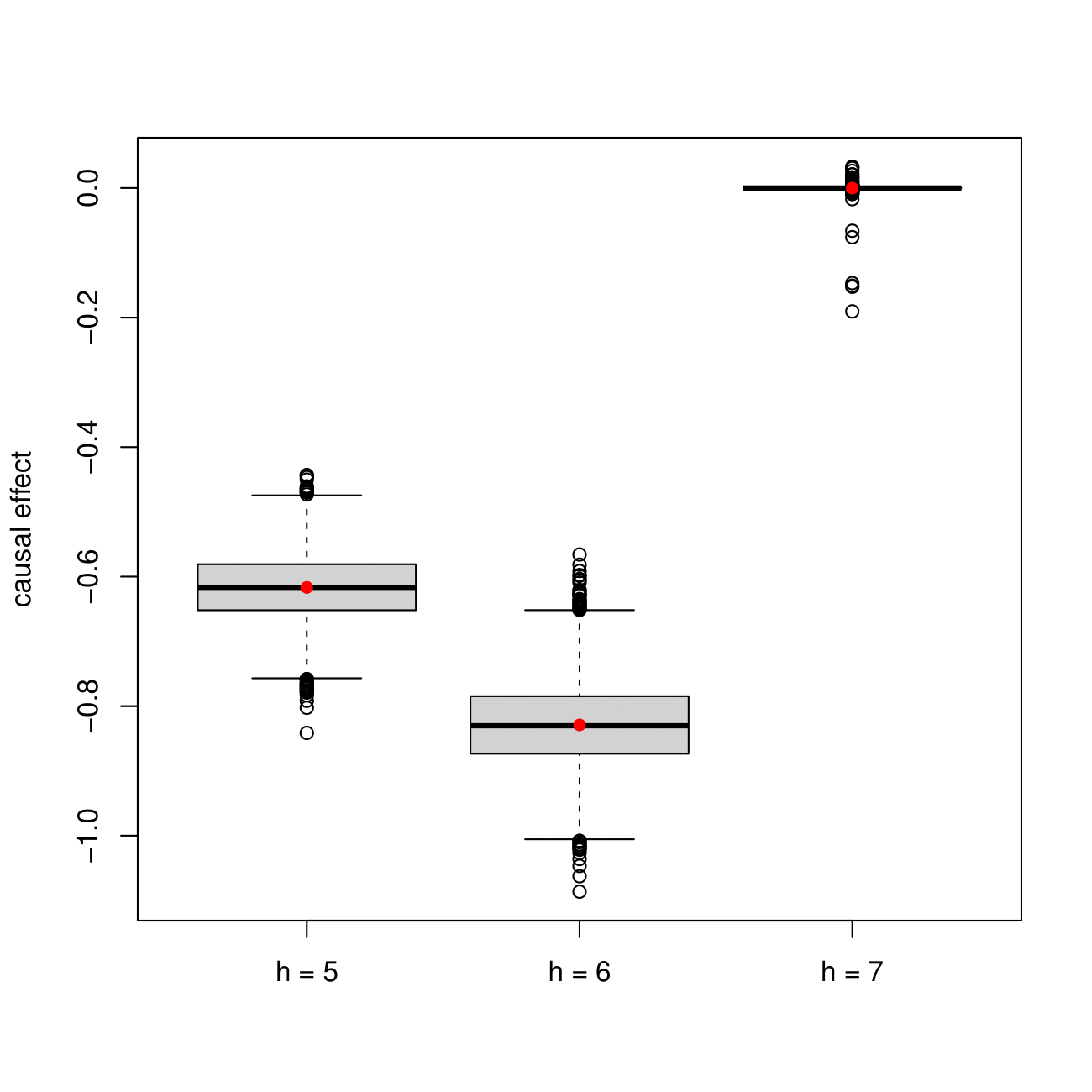}
	\caption{Simulated data. Posterior distribution of the causal effect coefficient $\theta^I_{h,1}$ for $h \in I$ and target of intervention $I = \{5,6,7\}$. BMA estimates are represented as red dots.}
	\label{fig:causal_effect}
\end{figure}

\texttt{joint\_causal} thus consists of a $(S,3)$ matrix, with each column referring to one of the three intervened variables. The posterior distribution of each causal effect coefficient $\theta_{h,Y}^I$, $h \in I$, is summarized in the box-plots of Figure \ref{fig:causal_effect}.

By setting \texttt{BMA = TRUE}, a Bayesian Model Averaging (BMA) estimate of the three parameters is returned. Each BMA estimate is obtained as the sample mean of the corresponding posterior draws in \texttt{joint\_causal}; see also Equation \eqref{eq:BMA:causal:effect}.

\begin{verbatim}
	out_causal_BMA <- get_causaleffect(learnDAG_output = out_mcmc,
	                  targets = c(5,6,7), response = 1, BMA = TRUE)
	                  
	round(out_causal_BMA, 4)
	
	  h = 5   h = 6   h = 7 
	-0.6167 -0.8290 -0.0001
\end{verbatim}

\subsection{Real data analyses}

In this section we consider a biological dataset of patients affected by Acute Myeloid Leukemia (AML).
The dataset contains the levels of $q = 18$ proteins and phosphoproteins involved in apoptosis and cell cycle regulation according to the KEGG database \citep{Kanehisa:et:al:2012} and is provided as a supplement of
\citet{Kornblau:et:al:2009}.
Measurements relative to $256$ newly diagnosed AML patients are included in the original dataset.
In addition, subjects are classified according to the French-American-British
(FAB) system
into several different AML subtypes. The same dataset was analyzed by \citet{Pete:etal:2015} and \citet{Castelletti:et:al:Stat:medicine:2020} from a multiple graphical model perspective to estimate group-specific dependence structures based on undirected and directed graphs respectively.

\begin{figure}[h]
	\centering
	\includegraphics[width=0.9\linewidth]{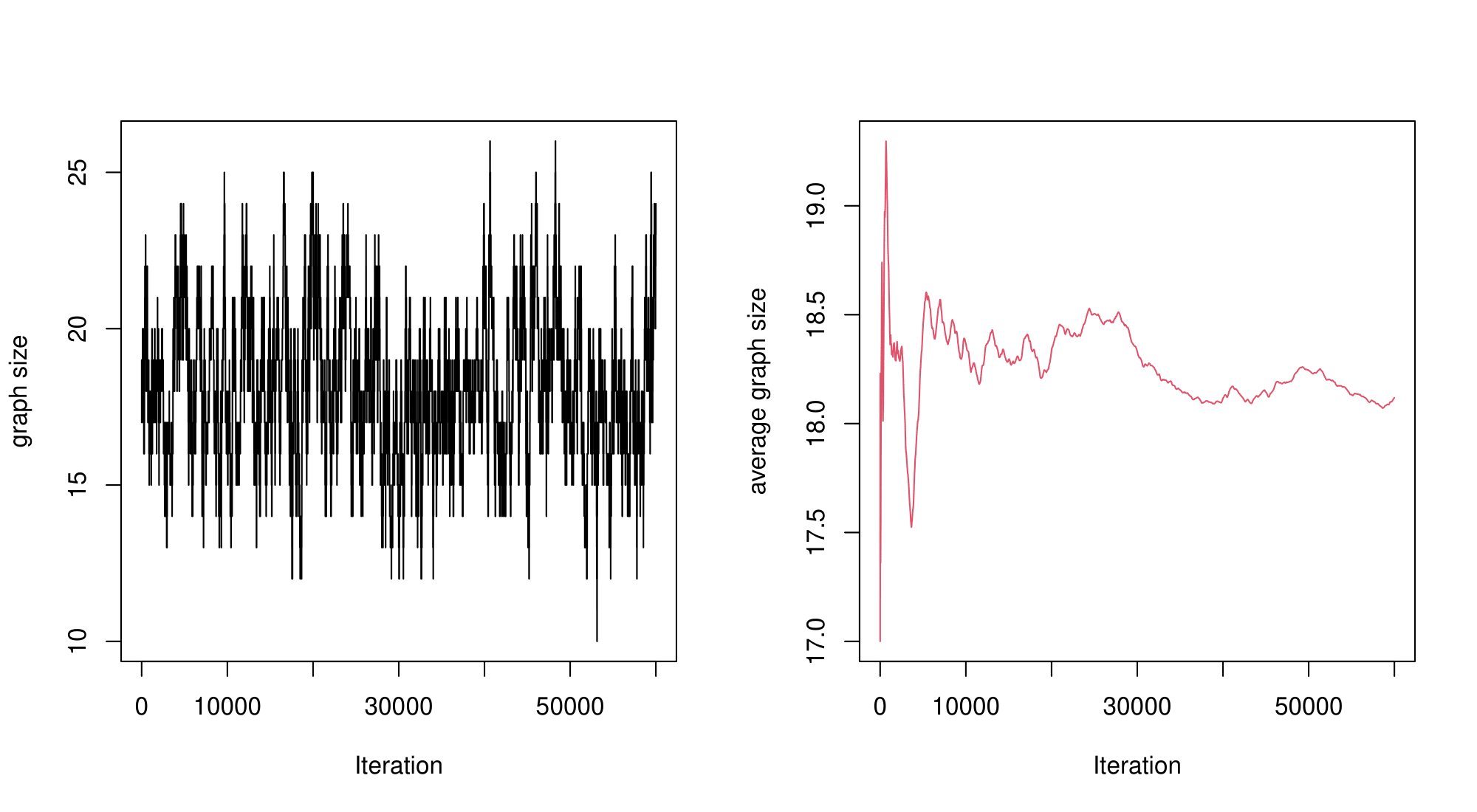}
	\caption{Leukemia data. Trace plot of the number of edges (graph size) in the DAGs visited by the MCMC at each iteration (left panel). Trace plot of the average number of edges in the DAGs computed up to iteration $s$, $s=1,\dots,60000$ (right panel).}
	\label{fig:edge:trace:plots}
\end{figure}

In the sequel we focus on subtype M2 and consider the $n=68$ corresponding observations which are collected in \texttt{leukemia\_data.csv}. We run the MCMC scheme in Algorithm \ref{alg:MCMC} and implemented by \texttt{learn\_DAG} by fixing
$S= 60000, B = 5000, a = q, \bU = \frac{1}{n}\bI_q$, $w = 0.5$:

\begin{verbatim}
	X <- read.csv("leukemia_data.csv", sep = ",")
	q <- ncol(X); n <- nrow(X)
	out_mcmc <- learn_DAG(S = 60000, burn = 5000, data = X, a = q,
	            U = diag(1,q)/n, w = 0.5, fast = TRUE, collapse = FALSE)
\end{verbatim}

We first perform MCMC diagnostics of convergence (Section \ref{sec:MCMC:diagnostics}) by running:
\begin{verbatim}
	get_diagnostics(learnDAG_output = out_mcmc)
\end{verbatim}

We report in the left-side panel of Figure \ref{fig:edge:trace:plots} the trace plot of the number of edges in the graphs visited by the MCMC at each iteration $s=1,\dots,S$.
The right-side panel of the same figure instead reports the trace plot of the \emph{average} number of edges in the DAGs visited by the MCMC up to iteration $s$, for $s = 1, \dots, S$. The apparent absence of trends in the trace plot and the curve stabilization around an average value both suggest a good degree of MCMC mixing and convergence to the target distribution.

\begin{figure}[h]
	\centering
	\includegraphics[width=0.95\linewidth]{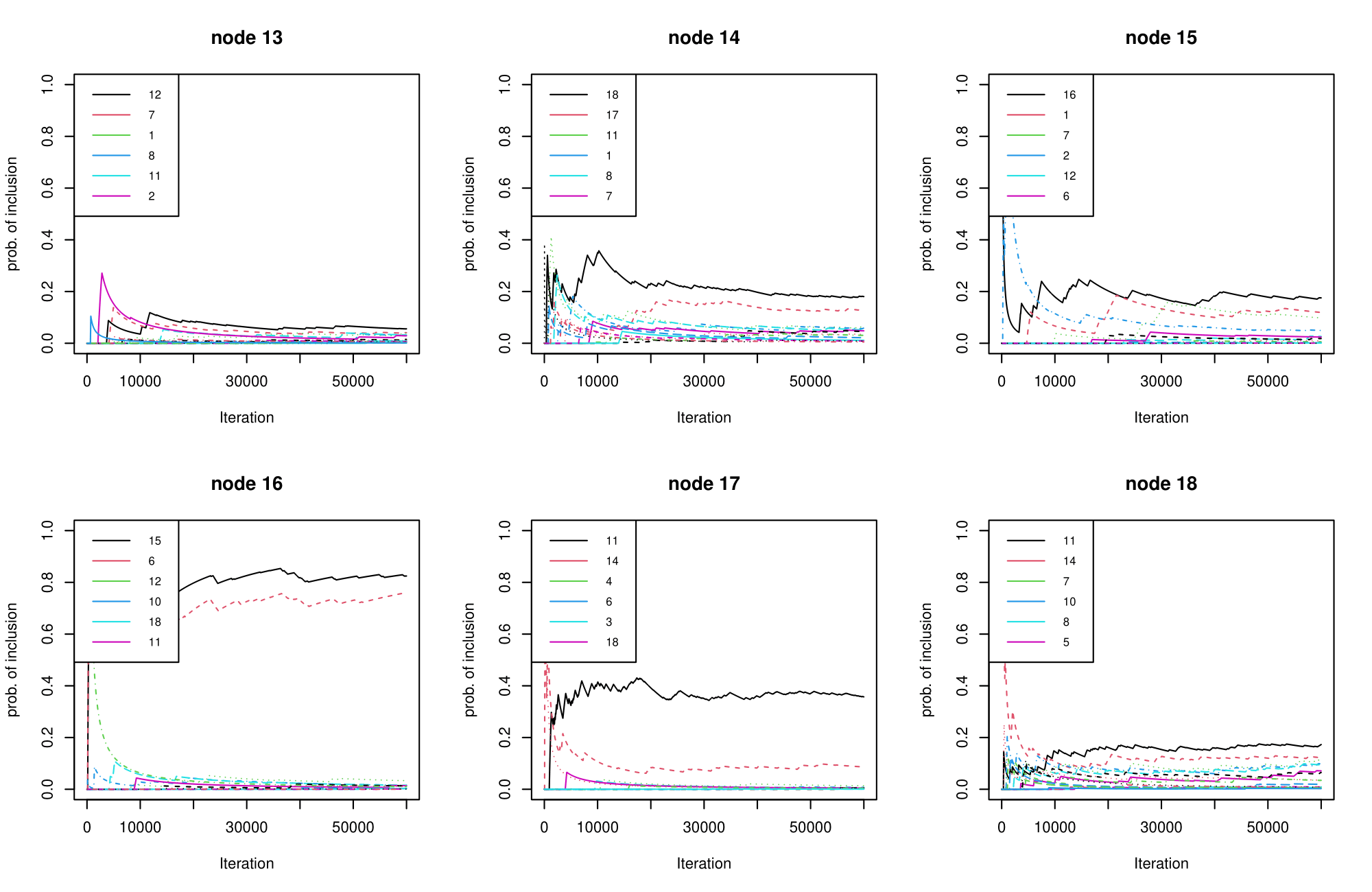}
	\caption{Leukemia data. Trace plot of the posterior probability of inclusion of $u \rightarrow v$, computed across iteration $s$, for selected nodes $v\in\{13,\dots,18\}$. For each node $v$, the six nodes $u$ associated with the highest posterior probabilities $p_{u\rightarrow v}$ are reported in the legend.}
	\label{fig:edge:probs}
\end{figure}

Figure \ref{fig:edge:probs} summarizes the behaviour of the posterior probabilities of inclusion of selected edges across MCMC iterations.
Specifically, each plot refers to one node $v\in\{13,\dots,18\}$ and contains a collection of trace plots each representing the posterior probability of inclusion of $u \rightarrow v$, ($u \ne v$) computed up to iteration $s$, for $s=1,\dots, S$.
By inspection, we can appreciate the convergence of each curve around stable values.

\begin{figure}[h]
	\centering
	\includegraphics[width=0.9\linewidth]{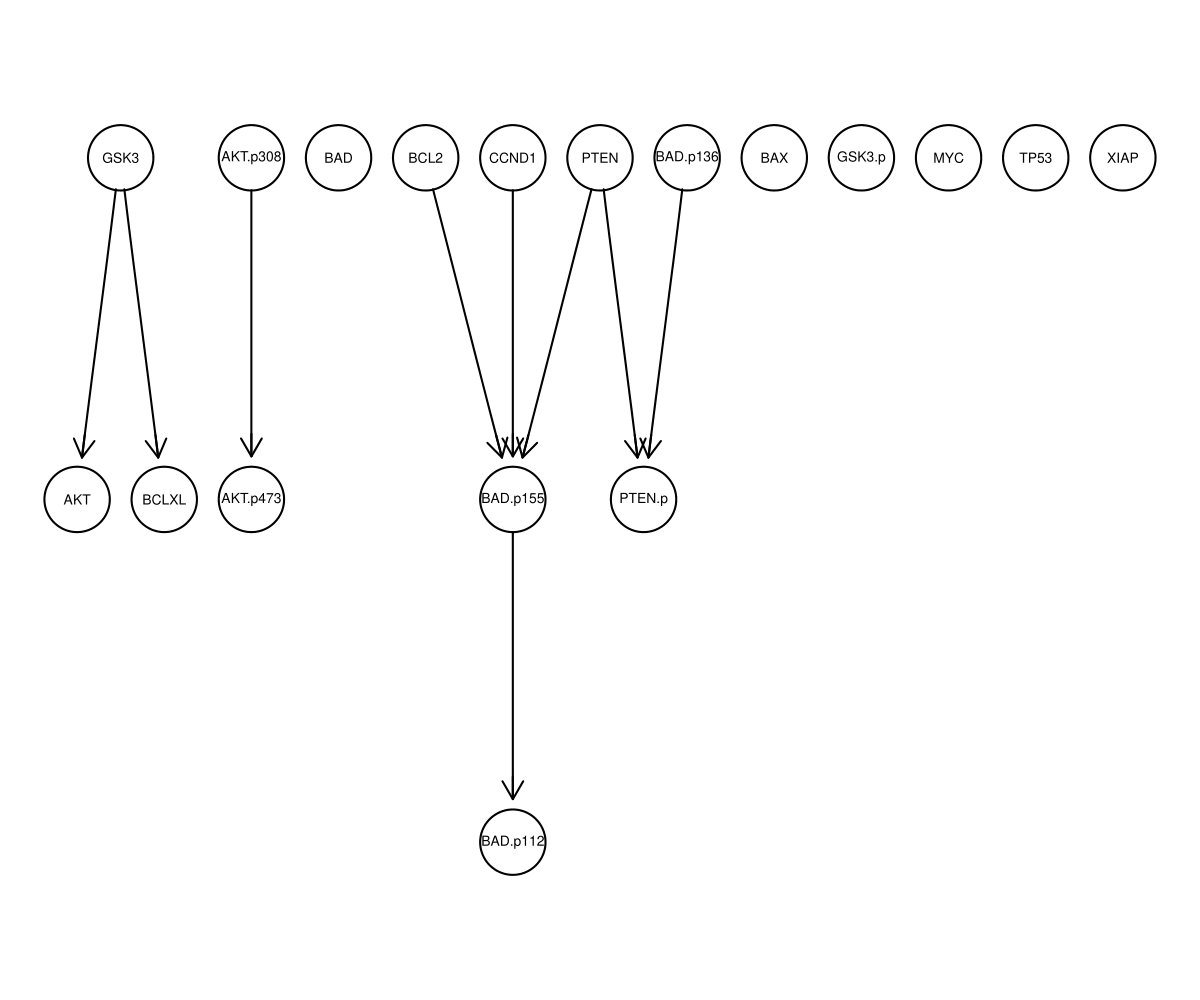}
	\caption{Leukemia data. Median Probability DAG-Model (MPM) estimate.}
	\label{fig:MPM:DAG}
\end{figure}

Given the MCMC output, we now proceed by constructing posterior summaries of interest.
Specifically, we first recover the posterior probabilities of inclusion $\hat p_{u\rightarrow v}$ in Equation \ref{eq:posterior:edge:inclusion} for each $u\rightarrow v$, $u \ne v$. These are collected in the $(q,q)$ matrix returned by \texttt{get\_edgeprobs} which is obtained as follows:
\begin{verbatim}
	post_edge_probs <- get_edgeprobs(learnDAG_output = out_mcmc)
\end{verbatim}

In Figure \ref{fig:edge:probs} we represent the resulting matrix \texttt{post\_edge\_probs} as a heat-map, with a dot at position $(u,v)$ corresponding to the estimated probability $\hat p_{u\rightarrow v}$.
In addition, we recover the Median Probability DAG-Model (MPM) as:

\begin{verbatim}
	MPM_dag <- get_MPMdag(out_mcmc)
\end{verbatim}

Object \texttt{MPM\_dag} corresponds to the adjacency matrix of the resulting graph estimate, which is represented in Figure \ref{fig:MPM:DAG}.

\begin{figure}[h]
	\centering
	\includegraphics[width=0.6\linewidth]{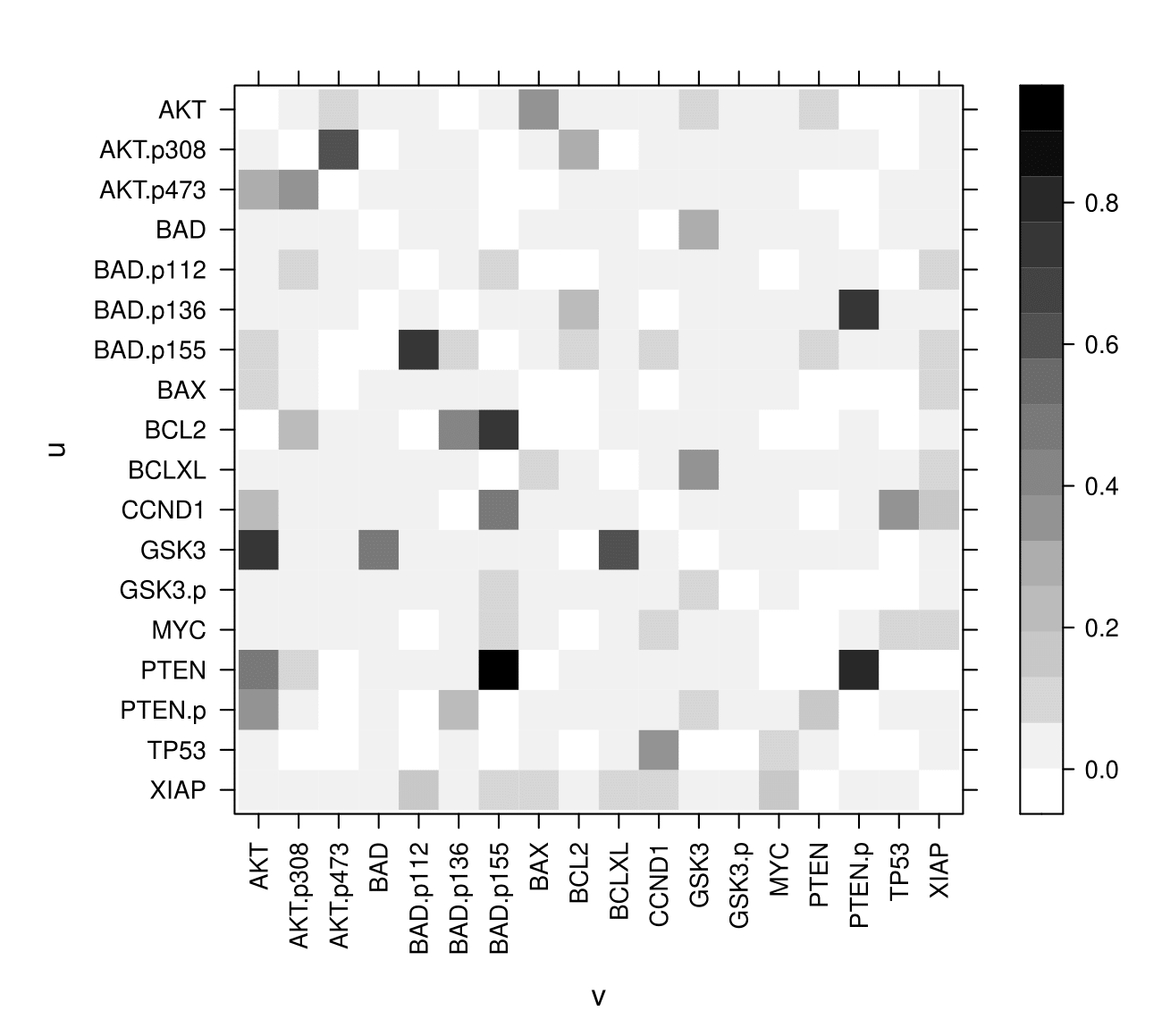}
	\caption{Leukemia data. Heat map with posterior probabilities of edge inclusion, for each directed edge $u \rightarrow v$.}
	\label{fig:heat_map}
\end{figure}

We now focus on causal effect estimation.
Among the proteins included in the study, AKT has received particular interest because of its role played in leukemogenesis. Specifically, AKT belongs to the PI3K-Akt-mTOR pathway, which is one of the intracellular pathways aberrantly up-regulated in AML; see also 
\citet{Nepstad:et:al:2020}.
Because of its role played in AML progression, we then consider the AKT protein
as the response of interest for our causal-effect analysis.

\begin{figure}[h]
	\centering
	\includegraphics[width=0.8\linewidth]{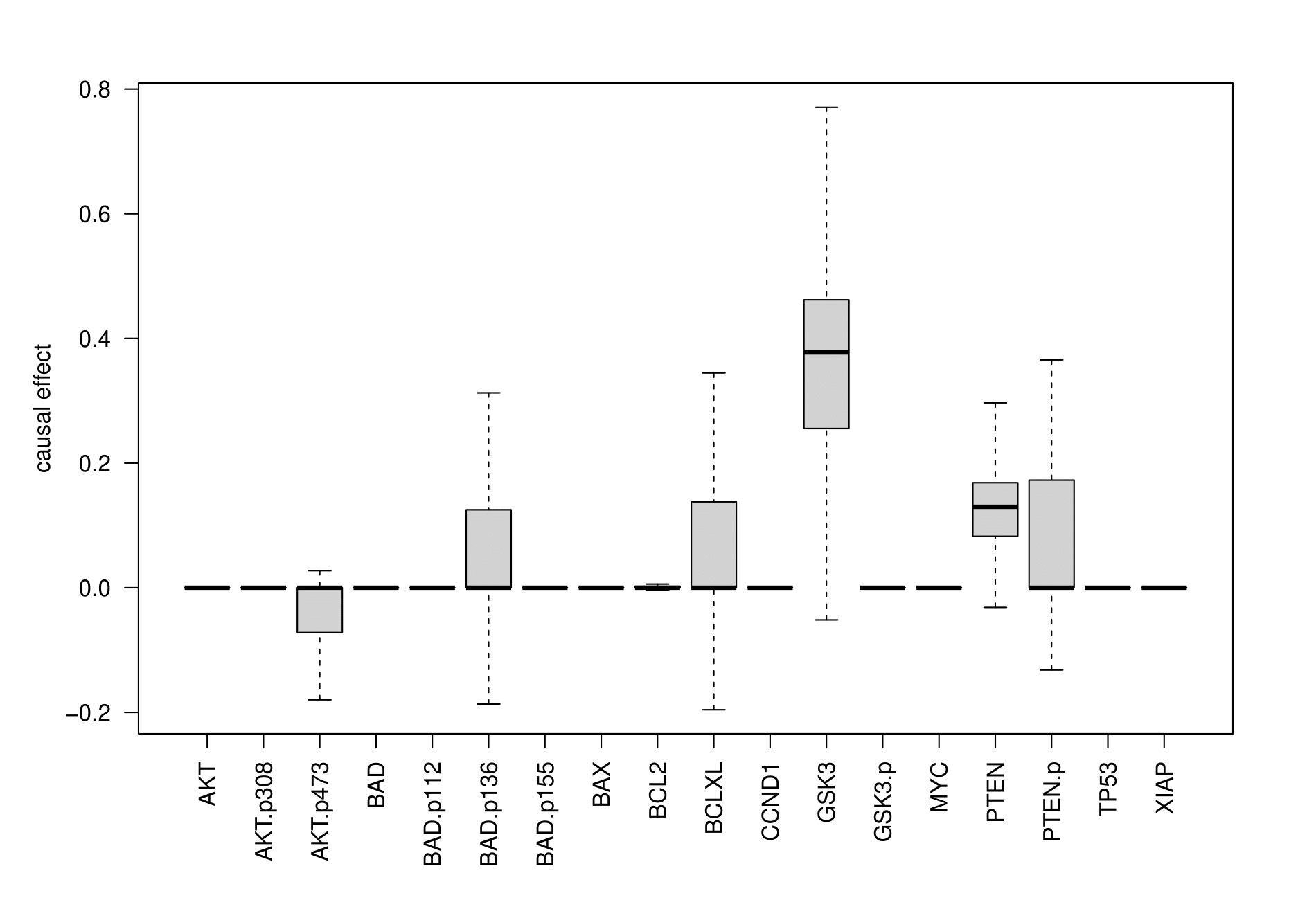}
	\caption{Leukemia data. Posterior distribution of the causal effect coefficient of an intervention $\text{do}\{X_j = \tilde{x}_j\}$ on response $Y=\textnormal{AKT}$, for $j\in\{1,\dots,q\}$ (each node labelled with the corresponding protein name).}
	\label{fig:leukemia:causal:effects}
\end{figure}

We first evaluate the causal effect of a single-node $(I=j)$ intervention $\text{do}\{X_j = \tilde{x}_j\}$ on $Y$ for each $j\in\{1,\dots,q\}$. The resulting collection of coefficients can provide useful information on the effect of selected interventions on each protein w.r.t.~AKT. In particular, this can help identifying promising targets for AKT regulation.
Results, in terms of posterior distributions of coefficients $\theta_{j,Y}^{j}$, for $j=1,\dots,q$, are obtained by running
\begin{verbatim}
	causal_all <- sapply(1:q, function(h) get_causaleffect(out_mcmc,
	              targets = j, response = 1, BMA = FALSE))
\end{verbatim}
and are summarized in the box-plots of Figure \ref{fig:leukemia:causal:effects}.

As a further example, consider a joint intervention acting on the GSK3 protein and phosphoprotein (GSK3 and GSK3.p respectively, corresponding to variables/columns 12 and 13 in the original data matrix \texttt{X}).
We evaluate the causal effect of such an intervention w.r.t.~AKT, by computing the BMA estimates of the two causal effect coefficients:

\begin{verbatim}
	causal_out_12_13 <- get_causaleffect(learnDAG_output = out_mcmc,
	                    targets = c(12,13), response = 1, BMA = F)
	round(causal_out_12_13, 4)
	h = 12 h = 13 
	0.3122 0.0050 
\end{verbatim}

\section{Conclusions}
\label{sec:discussion}

In this article we introduced \textbf{\texttt{BCDAG}}, a novel \textbf{\texttt{R}} package for Bayesian structure learning and causal effect estimation from Gaussian, observational data. This package implements the method of \citet{Castelletti:Mascaro:2021} through a variety of functions that allow
for DAG-model selection, posterior inference of model parameters and estimation of causal effects.
The proposed implementation scales efficiently to an arbitrarily large number of observations and, whenever the visited DAGs are sufficiently \textit{sparse}, to datasets with a large number of variables.
However, convergence of MCMC algorithms in high dimensional (large $q$) settings can be extremely slow, thus hindering posterior inference. For this reason, we plan to extend the proposed method by considering alternative MCMC schemes that may improve convergence to the target posterior distribution; see for instance \citet{Kuipers:et:al:2021:arxiv} and \citet{Agrawal:2018:pmlr}. In addition, we are considering extensions of the proposed structure learning method to more general settings where a combination of observational and interventional (experimental) data is available
\citep{Castelletti:Consonni:2019:AOAS}.

We finally encourage \textbf{\texttt{R}} users to provide feedback, questions or suggestion by e-mail correspondence or directly through our \texttt{github} code repository \url{https://github.com/alesmascaro/BCDAG}.

\bibliographystyle{biometrika} 
\bibliography{biblio}

\end{document}